\documentclass[11pt,a4paper]{article}
\usepackage[hyperref]{emnlp-ijcnlp-2019}
\usepackage{times}
\usepackage{latexsym}
\usepackage{multirow}
\usepackage{amssymb, amsmath}
\usepackage{url}
\usepackage{graphicx}
\usepackage{booktabs}
\usepackage{todonotes}
\usepackage{fontawesome}

\aclfinalcopy 

\title{Findings of the NLP4IF-2019 Shared Task\\ on Fine-Grained Propaganda Detection}

\author{\textrm{Giovanni Da San Martino}$^{1}$ \quad Alberto Barr\'{o}n-Cede\~{n}o$^{2}$  \quad \textbf{Preslav Nakov}$^{1}$ \\
$^{1}$ Qatar Computing Research Institute, HBKU, Qatar \\   
$^{2}$ Universit\`{a} di Bologna, Forl\`{i}, Italy\\
{\tt \{gmartino, pnakov\}@hbku.edu.qa}\hspace{5mm}
{\tt a.barron@unibo.it}}

\date{}

\begin{document}

\maketitle
\begin{abstract}
We present the shared task on Fine-Grained Propaganda Detection, which was organized as part of the NLP4IF workshop at EMNLP-IJCNLP 2019. There were two subtasks.  FLC is a fragment-level task that asks for the identification of propagandist text fragments in a news article and also for the prediction of the specific propaganda technique used in each such fragment (18-way classification task). SLC is a sentence-level binary classification task asking to detect the sentences that contain propaganda.
A total of 12 teams submitted systems for the FLC task, 25 teams did so for the SLC task, and 14 teams eventually submitted a system description paper. For both subtasks, most systems managed to beat the baseline by a sizable margin. The leaderboard and the data from the competition are available at \url{http://propaganda.qcri.org/nlp4if-shared-task/}.
\end{abstract}

\section{Introduction}

Propaganda aims at influencing people's mindset with the purpose of advancing a specific agenda. 
In the Internet era, thanks to the mechanism of sharing in social networks, propaganda campaigns have the potential of reaching very large audiences~\cite{Glowacki:18,Muller2018,brazil}. 

Propagandist news articles use specific techniques to convey their message, such as \textit{whataboutism}, \textit{red Herring}, and \textit{name calling}, among many others (cf.\ Section~\ref{sec:background}). 
Whereas proving intent is not easy, we can analyse the language of a claim/article and look for the use of specific propaganda techniques. Going at this fine-grained level can yield more reliable systems and it also makes it possible to explain to the user why an article was judged as propagandist by an automatic system.

\noindent With this in mind, we organised the shared task on fine-grained propaganda detection at the NLP4IF@EMNLP-IJCNLP 2019 workshop. 
The task is based on a corpus of news articles annotated with an inventory of 18 propagandist techniques at the fragment level. 
We hope that the corpus would raise interest outside of the community of researchers studying propaganda. For example, the techniques related to fallacies and the ones relying on emotions might provide a novel setting for researchers interested in Argumentation and Sentiment Analysis.

\section{Related Work}
\label{sec:relatedwork}

Propaganda has been tackled mostly at the article level. \citet{rashkin-EtAl:2017:EMNLP2017} created a corpus of news articles labelled as propaganda, trusted, hoax, or satire. 
\citet{AAAI2019:proppy} experimented with a binarized version of that corpus: propaganda vs. the other three categories.
\citet{BARRONCEDENO20191849} annotated a large binary corpus of propagandist vs. non-propagandist articles and proposed a feature-based system for discriminating between them. 
In all these cases, the labels were obtained using distant supervision, assuming that all articles from a given news outlet share the label of that outlet, which inevitably introduces noise~\cite{Horne2018}. 

A related field is that of computational argumentation which, among others, deals with some logical fallacies related to propaganda. 
\citet{Habernal2018} presented a corpus of Web forum discussions with instances of \textit{ad hominem} fallacy. 
\citet{Habernal.et.al.2017.EMNLP,Habernal2018b} introduced \textit{Argotario}, a game to educate people to recognize and create fallacies, a by-product of which is a corpus with $1.3k$ arguments annotated with five fallacies such as \textit{ad hominem}, \textit{red herring} and \textit{irrelevant authority}, which directly relate to propaganda.

\noindent Unlike~\cite{Habernal.et.al.2017.EMNLP,Habernal2018b,Habernal2018}, our corpus uses 18 techniques annotated on the same set of news articles. Moreover, our annotations aim at identifying the minimal fragments related to a technique instead of flagging entire arguments. 

The most relevant related work is our own, which is published in parallel to this paper at EMNLP-IJCNLP 2019~\cite{EMNLP19DaSanMartino} and describes a corpus that is a subset of the one used for this shared task.

\section{Propaganda Techniques}
\label{sec:background}

Propaganda uses psychological and rhetorical techniques to achieve its objective. Such techniques include the use of logical fallacies and appeal to emotions.
For the shared task, we use 18 techniques that can be found in news articles and can be judged intrinsically, without the need to retrieve supporting information from external resources. We refer the reader to \cite{EMNLP19DaSanMartino} for more details on the propaganda techniques; below we report the list of techniques:

\paragraph{1. Loaded language.}
Using words/phrases with strong emotional implications (positive or negative) to influence an audience~\cite[p.~6]{Weston2000}.

\paragraph{2. Name calling or labeling.}
Labeling the object of the propaganda as something the target audience fears, hates, finds undesirable or otherwise loves or praises~\cite{Miller}.

\paragraph{3. Repetition.} 
Repeating the same message over and over again, so that the audience will eventually accept it~\cite{Torok2015,Miller}.

\paragraph{4. Exaggeration or minimization.}
Either representing something in an excessive manner: making things larger, better, worse, or making something seem less important or smaller than it actually is~\cite[p.~303]{Jowett2012a}, e.g.,~saying that an insult was just a joke.

\paragraph{5. Doubt.}
Questioning the credibility of someone or something.

\paragraph{6. Appeal to fear/prejudice.} 
Seeking to build support for an idea by instilling anxiety and/or panic in the population towards an alternative, possibly based on preconceived judgments.

\paragraph{7. Flag-waving.} 
Playing on strong national feeling (or with respect to a group, e.g.,~race, gender, political preference) to justify or promote an action or idea~\cite{Hobbs2008}.

\paragraph{8. Causal oversimplification.} Assuming one cause when there are multiple causes behind an issue. 
We include \emph{scapegoating} as well: the transfer of the blame to one person or group of people without investigating the complexities of an issue. 

\paragraph{9. Slogans.}
A brief and striking phrase that may include labeling and stereotyping. Slogans tend to act as emotional appeals~\cite{As2015}.

\paragraph{10. Appeal to authority.}
Stating that a claim is true simply because a valid authority/expert on the issue supports it, without any other supporting evidence~\cite{Goodwin2011}. We include the special case where the reference is not an authority/expert, although it is referred to as \emph{testimonial} in the literature~\cite[p.~237]{Jowett2012a}.

\paragraph{11. Black-and-white fallacy, dictatorship.}
Presenting two alternative options as the only possibilities, when in fact more possibilities exist \cite{Torok2015}. As an extreme case, telling the audience exactly what actions to take, eliminating any other possible choice (\emph{dictatorship}).

\paragraph{12. Thought-terminating \textit{clich\'e}.}
Words or phrases that discourage critical thought and meaningful discussion about a given topic. They are typically short and generic sentences that offer seemingly simple answers to complex questions or that distract attention away from other lines of thought~\cite[p.~78]{Hunter2015}.

\paragraph{13. Whataboutism.} Discredit an opponent's position by charging them with hypocrisy without directly disproving their argument~\cite{Richter2017}. 

\paragraph{14. Reductio ad Hitlerum.} 
Persuading an audience to disapprove an action or idea by suggesting that the idea is popular with groups hated in contempt by the target audience. It can refer to any person or concept with a negative connotation~\cite{Aper2009}.

\begin{figure*}[t]
    \centering
	\includegraphics[width=\textwidth]{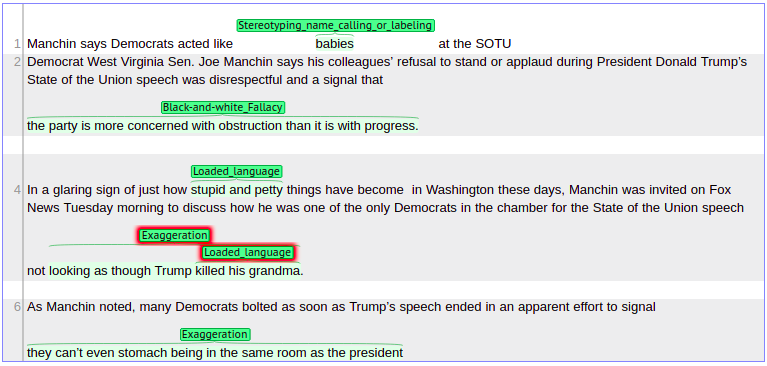}
    \caption{The beginning of an article with annotations.\label{fig:example}}
\end{figure*}

\paragraph{15. Red herring.}
Introducing irrelevant material to the issue being discussed, so that everyone's attention is diverted away from the points made~\cite[p.\ 78]{Weston2000}. 
Those subjected to a red herring argument are led away from the issue that had been the focus of the discussion and urged to follow an observation or claim that may be associated with the original claim, but is not highly relevant to the issue in dispute~\cite{Aper2009}.

\paragraph{16. Bandwagon.}
Attempting to persuade the target audience to join in and take the course of action because ``everyone else is taking the same action''~\cite{Hobbs2008}.

\paragraph{17. Obfuscation, intentional vagueness, confusion.} 
Using deliberately unclear words, to let the audience have its own interpretation~\cite[p.~8]{Suprabandari2007,Weston2000}. 
For instance, when an unclear phrase with multiple possible meanings is used within the argument and, therefore, it does not really support the conclusion.

\paragraph{18. Straw man.} When an opponent's proposition is substituted with a similar one which is then refuted in place of the original~\cite{Walton1996}.

\section{Tasks}
\label{sec:tasks}

The shared task features two subtasks: 

\paragraph{Fragment-Level Classification task (FLC).} Given a news article, detect all spans of the text in which a propaganda technique is used. In addition, for each span the propaganda technique applied must be identified. 

\paragraph{Sentence-Level Classification task (SLC).} A sentence is considered propagandist if it contains at least one propagandist fragment. 
We then define a binary classification task in which, given a sentence, the correct label, either \emph{propaganda} or \emph{non-propaganda}, is to be predicted.


\section{Data}
\label{sec:data}

The input for both tasks consists of news articles in free-text format, collected from 36 propagandist and 12 non-propagandist news outlets\footnote{We obtained the gold labels about whether a given news outlet was propagandistic from the Media Bias Fact Check website: \url{http://mediabiasfactcheck.com/}} and then annotated by professional annotators. More details about the data collection and the annotation, as well as statistics about the corpus can be found in~\cite{EMNLP19DaSanMartino}, where an earlier version of the corpus is described, which includes 450 news articles. We further annotated 47 additional articles for the purpose of the shared task using the same protocol and the same annotators.

The training, the development, and the test partitions of the corpus used for the shared task consist of 350, 61, and 86 articles and of 16,965, 2,235, and 3,526 sentences, respectively. 
Figure~\ref{fig:example} shows an annotated example, which contains several propaganda techniques. For example, the fragment \emph{babies} on line 1 is an instance of both \texttt{Name\_Calling} and \texttt{Labeling}. 
Note that the fragment \emph{not looking as though Trump killed his grandma} on line 4 is an instance of \texttt{Exaggeration\_or\_Minimisation} and it overlaps with the fragment \emph{killed his grandma}, which is an instance of \texttt{Loaded\_Language}.

Table~\ref{tab:annstats} reports the total number of instances per technique and the percentage with respect to the total number of annotations, for the training and for the development sets.

\begin{table}
    \centering
    \small
    \begin{tabular}{l@{\hspace{1mm}}r@{\hspace{0.5mm}}r   @{\hspace{4mm}}r@{\hspace{0.5mm}}r}
    \toprule    
\textbf{Technique}          & \bf Train & \bf (\%)\,\,\,& \bf Dev & \bf (\%)\,\,\,  \\
\midrule
Appeal to Authority         &  116  & (1.92) & 50   &    (5.92)\\
Appeal to fear / prejudice  &  239  & (3.96) & 103   & (12.19) \\
Bandwagon                   &   13  & (0.22) & 3   & (0.36)\\
Black and White Fallacy     &  109  & (1.80) & 17   & (2.01)\\
Causal Oversimplification   &  201  & (3.33) & 22   & (2.60)\\
Doubt                       &  490  & (8.11) & 39   & (4.62)\\
Exaggeration, Minimisation  &  479  & (7.93) & 59   & (6.98)\\
Flag Waving                 &  240  & (3.97) & 63   & (7.46)\\
Loaded Language             & 2,115 & (35.10) & 229   & (27.10)\\
Name Calling, Labeling      & 1,085 & (17.96) & 87   & (10.30)\\
Obfuscation, Intentional \\ ~~~Vagueness, Confusion 
                            &   11  & (0.18) & 5   & (0.59)\\
Red Herring                 &   33  & (0.55) & 10   & (1.18)\\
Reductio ad hitlerum        &   54  & (0.89) & 9   & (1.07)\\
Repetition                  &  571  & (9.45) &101   & (11.95)\\
Slogans                     &  136  & (2.25) & 26   & (3.08)\\
Straw Men                   &   13  & (0.22) & 2   & (0.24)\\
Thought-terminating Cliches &   79  & (1.31) & 10   & (1.18)\\
Whataboutism                &   57  & (0.94) & 10   & (1.18)\\
\bottomrule
    \end{tabular}
    \caption{Statistics about the gold annotations for the training and the development sets. \label{tab:annstats}}
\end{table}


\section{Setup}
\label{sec:setup}

The shared task had two phases: In the development phase, the participants were provided labeled training and development datasets; in the testing phase, testing input was further provided.

\begin{description}
\item[Phase 1.] The participants tried to achieve the best performance on the development set. A live leaderboard kept track of the submissions.
\item[Phase 2.] The test set was released and the participants had few days to make final predictions. \end{description}
In phase~2, no immediate feedback on the submissions was provided. The winner was determined based on the performance on the test set.


\section{Evaluation}
\label{sec:evaluation}

\paragraph{FLC task.}
FLC is a composition of two subtasks: the identification of the propagandist text fragments and the identification of the techniques used (18-way classification task). While F$_1$ measure is appropriate for a multi-class classification task, we modified it to account for partial matching between the spans; see \cite{EMNLP19DaSanMartino} for more details. We further computed an F$_1$ value for each propaganda technique (not shown below for the sake of saving space, but available on the leaderboard).

\paragraph{SLC task.}
SLC is a binary classification task with imbalanced data. Therefore, the official evaluation measure for the task is the standard F$_1$ measure. We further report Precision and Recall.


\section{Baselines\label{sec:baselines}}

The baseline system for the SLC task is a very simple logistic regression classifier with default parameters, where we represent the input instances with a single feature: the length of the sentence. The performance of this baseline on the SLC task is shown in Tables~\ref{tab:slctest} and~\ref{tab:slcdev}. 

The baseline for the FLC task generates spans and selects one of the 18 techniques randomly. The inefficacy of such a simple random baseline is illustrated in Tables~\ref{tab:flc:results} and~\ref{tab:flc:dev}.

\section{Participants and Approaches}
\label{sec:approaches}

A total of 90 teams registered for the shared task, and 39 of them submitted predictions for a total of 3,065 submissions. For the FLC task, 21 teams made a total of 527 submissions, and for the SLC task, 35 teams made a total of 2,538 submissions. 

\begin{table*}
\centering
\small
\begin{tabular}{lcccccccc}
\toprule
\bf Team  & BERT	    & LSTM  		& Word Emb.  & Char. Emb.  	  & Features  	   & Unsup. Tuning \\
\midrule
CUNLP 
&       	    & \faCheckSquare    & \faCheckSquare  & \faCheckSquare &	\\
Stalin 
& \faCheckSquare  & \faCheckSquare	\\
MIC-CIS 
 & \faCheckSquare  &  \faCheckSquare   &          	  &           	   & \faCheckSquare      &           \\
ltuorp 
	  & \faCheckSquare  & 			& 	   	  & 		   & 		& \\ 
ProperGander 
& \faCheckSquare & \faCheckSquare	\\
newspeak 
 & \faCheckSquare  &           	&           	  &           	   &           & \faCheckSquare    \\
\bottomrule
\end{tabular}
\caption{Overview of the approaches for the fragment-level classification task.\label{tab:flc}}
\end{table*}

\begin{table*}
\centering
\small
\begin{tabular}{lcccccccc}
\toprule
\bf Team  & BERT	    & LSTM		& logreg  	 & USE & CNN 	& Embeddings	& Features 	& Context \\
\midrule
NSIT 
& \faCheckSquare  & \faCheckSquare	\\
CUNLP 
& \faCheckSquare  & 			& \faCheckSquare &     &	&             	& \faCheckSquare    & \\
JUSTDeep & \faCheckSquare  & \faCheckSquare	& 		 &     &	& \faCheckSquare& \faCheckSquare 	\\
Tha3aroon
& \faCheckSquare  & 			&		 & \faCheckSquare\\			
LIACC	  & 		    & \faCheckSquare 	&	    	 &     &	& \faCheckSquare & \faCheckSquare	\\
MIC-CIS 
& \faCheckSquare  & 			& \faCheckSquare &     &\faCheckSquare  &  \faCheckSquare       & \faCheckSquare    & \\
CAUnLP     & \faCheckSquare  & 			&         	 &     &	&             	&         	& \faCheckSquare  \\
YMJA 
  & \faCheckSquare  \\
jinfen 
  & \faCheckSquare  & 			& \faCheckSquare &     &	& 		& \faCheckSquare	\\
ProperGander 
& \faCheckSquare  \\
\bottomrule
\end{tabular}
\caption{\label{tab:slc}Overview of the approaches used for the sentence-level classification task.}
\end{table*}
Below, we give an overview of the approaches as described in the participants' papers. 
Tables~\ref{tab:flc} and~\ref{tab:slc} offer a high-level summary.

\subsection{Teams Participating in the Fragment-Level Classification Only}

Team \textbf{newspeak}~\cite{Yoosuf:19} achieved the best results on the test set for the FLC task using 20-way word-level classification based on BERT~\cite{devlin2018bert}: a word could belong to one of the 18 propaganda techniques, to none of them, or to an auxiliary (token-derived) class. The team fed one sentence at a time in order to reduce the workload. In addition to experimenting with an out-of-the-box BERT, they also tried unsupervised fine-tuning both on the 1M news dataset and on Wikipedia. Their best model was based on the uncased base model of BERT, with 12 Transformer layers~\cite{DBLP:conf/nips/VaswaniSPUJGKP17}, and 110 million parameters. Moreover, oversampling of the least represented classes proved to be crucial for the final performance. Finally, careful analysis has shown that the model pays special attention to adjectives and adverbs.

Team \textbf{Stalin}~\cite{Ek:19} focused on data augmentation to address the relatively small size of the data for fine-tuning contextual embedding representations based on ELMo~\cite{peters-EtAl:2018:N18-1}, BERT, and Grover~\cite{DBLP:journals/corr/abs-1905-12616}. The balancing of the embedding space was carried out by means of synthetic minority class over-sampling. Then, the learned representations were fed into an LSTM.

\subsection{Teams Participating in the Sentence-Level Classification Only}

Team \textbf{CAUnLP}~\cite{Hou:19} used two context-aware representations based on BERT. In the first representation, the target sentence is followed by the title of the article. In the second representation, the previous sentence is also added. They performed subsampling in order to deal with class imbalance, and experimented with BERT$_{BASE}$ and BERT$_{LARGE}$

Team \textbf{LIACC}~\cite{Ferreira:19} used hand-crafted features and pre-trained ELMo embeddings. They also observed a boost in performance when balancing the dataset by dropping some negative examples. 

Team \textbf{JUSTDeep}~\cite{AlOmari:19} used a combination of models and features, including word embeddings based on GloVe \cite{pennington-etal-2014-glove} concatenated with vectors representing affection and lexical features. These were combined in an ensemble of supervised models: bi-LSTM, XGBoost, and variations of BERT.

Team \textbf{YMJA}~\cite{Hua:19} also based their approach on fine-tuned BERT. Inspired by \textit{kaggle} competitions on sentiment analysis, they created an ensemble of models via cross-validation.

Team \textbf{jinfen}~\cite{Li:19} used a logistic regression model fed with a manifold of representations, including TF.IDF and BERT vectors, as well as vocabularies and readability measures.

Team \textbf{Tha3aroon}~\cite{Fadel:19} implemented an ensemble of three classifiers: two based on BERT and one based on a universal sentence encoder~\cite{DBLP:journals/corr/abs-1803-11175}. 

Team \textbf{NSIT}~\cite{Aggarwal:19} explored three of the most popular transfer learning models: various versions of ELMo, BERT, and
RoBERTa~\cite{roberta:2019}. 

Team \textbf{Mindcoders}~\cite{Vlad:19} combined BERT, Bi-LSTM and Capsule networks~\cite{DBLP:conf/nips/SabourFH17} into a single deep neural network and pre-trained the resulting network on corpora used for related tasks, e.g.,~emotion classification. 

Finally, team \textbf{ltuorp}~\cite{Mapes:19} used an attention transformer using BERT trained on Wikipedia and BookCorpus.

\subsection{Teams Participating in Both Tasks}

Team \textbf{MIC-CIS}~\cite{Gupta:19} participated in both tasks. For the sentence-level classification, they used a voting ensemble including logistic regression, convolutional neural networks, and BERT, in all cases using FastText embeddings \cite{bojanowski-etal-2017-enriching} and pre-trained BERT models. Beside these representations, multiple features of readability, sentiment and emotions were considered. For the fragment-level task, they used a multi-task neural sequence tagger, based on LSTM-CRF~\cite{DBLP:journals/corr/HuangXY15}, in conjunction with linguistic features. Finally, they applied sentence- and fragment-level models jointly.


\begin{table}[t]
	\centering
	\small
	\begin{tabular}{r@{\hspace{1mm}}p{2.2cm}ccc}
	\toprule	
	\multicolumn{5}{c}{\textbf{SLC Task: Test Set (Official Results)}}\\
\textbf{Rank} & \textbf{Team} & \textbf{F$_1$} & \textbf{Precision} & \textbf{Recall} \\
\midrule
\hfil 1 & \textbf{ltuorp} & \textbf{0.6323} & 0.6028 & 0.6648 \\
\hfil 2 & ProperGander & 0.6256 & 0.5649 & 0.7009 \\
\hfil 3 & YMJA & 0.6249 & 0.6252 & 0.6246 \\
\hfil 4 & MIC-CIS & 0.6230 & 0.5735 & 0.6818 \\
\hfil 5 & TeamOne & 0.6183 & 0.5778 & 0.6648 \\
\hfil 6 & Tha3aroon & 0.6138 & 0.5309 & 0.7274 \\
\hfil 7 & JUSTDeep & 0.6112 & 0.5792 & 0.6468 \\
\hfil 8 & CAUnLP & 0.6109 & 0.5180 & 0.7444 \\
\hfil 9 & LIPN & 0.5962 & 0.5241 & 0.6914 \\
\hfil 10 & LIACC & 0.5949 & 0.5090 & 0.7158 \\
\hfil 11 & aschern & 0.5923 & 0.6050 & 0.5800 \\
\hfil 12 & MindCoders & 0.5868 & 0.5995 & 0.5747 \\
\hfil 13 & jinfen & 0.5770 & 0.5059 & 0.6712 \\
\hfil 14 & guanggong & 0.5768 & 0.5039 & 0.6744 \\
\hfil 15 & Stano & 0.5619 & 0.6666 & 0.4856 \\
\hfil 16 & nlpseattle & 0.5610 & 0.6250 & 0.5090 \\
\hfil 17 & gw2018 & 0.5440 & 0.4333 & 0.7306 \\
\hfil 18 & SDS & 0.5171 & 0.6268 & 0.4400 \\
\hfil 19 & BananasInPajamas & 0.5080 & 0.5768 & 0.4538 \\
\hfil 20 & \underline{Baseline} & 0.4347 & 0.3880 & 0.4941 \\
\hfil 21 & NSIT & 0.4343 & 0.5000 & 0.3838 \\
\hfil 22 & Stalin & 0.4332 & 0.6696 & 0.3202 \\
\hfil 23 & Antiganda & 0.3967 & 0.6459 & 0.2863 \\
\hfil 24 & Debunkers & 0.2307 & 0.3994 & 0.1622 \\
\hfil 25 & SBnLP & 0.1831 & 0.2220 & 0.1558 \\
\hfil 26 & Sberiboba & 0.1167 & 0.5980 & 0.0646	\\
\bottomrule
	\end{tabular}
	\caption{Official test results for the SLC task.\label{tab:slctest}}
\end{table}

\begin{table}[t]
	\centering
	\small
	\begin{tabular}{r@{\hspace{1mm}}p{2.2cm}ccc}
	\toprule	
	\multicolumn{5}{c}{\textbf{SLC Task: Development Set}}\\
\textbf{Rank} & \textbf{Team} & \textbf{F$_1$} & \textbf{Precision} & \textbf{Recall} \\
\midrule
\hfil 1 & Tha3aroon & 0.6883 & 0.6104 & 0.7889 \\
\hfil 2 & KS & 0.6799 & 0.5989 & 0.7861 \\
\hfil 3 & CAUnLP & 0.6794 & 0.5943 & 0.7929 \\
\hfil 4 & ProperGander & 0.6767 & 0.5774 & 0.8173 \\
\hfil 5 & JUSTDeep & 0.6745 & 0.6234 & 0.7347 \\
\hfil 6 & ltuorp & 0.6700 & 0.6351 & 0.7090 \\
\hfil 7 & TeamOne & 0.6649 & 0.6198 & 0.7171 \\
\hfil 8 & aschern & 0.6646 & 0.6104 & 0.7293 \\
\hfil 9 & jinfen & 0.6616 & 0.5800 & 0.7699 \\
\hfil 10 & YMJA & 0.6601 & 0.6338 & 0.6887 \\
\hfil 11 & SBnLP & 0.6548 & 0.5674 & 0.7740 \\
\hfil 12 & guanggong & 0.6510 & 0.5737 & 0.7523 \\
\hfil 13 & LIPN & 0.6484 & 0.5889 & 0.7212 \\
\hfil 14 & Stalin & 0.6377 & 0.5957 & 0.6860 \\
\hfil 15 & Stano & 0.6374 & 0.6561 & 0.6197 \\
\hfil 16 & BananasInPajamas & 0.6276 & 0.5204 & 0.7902 \\
\hfil 17 & Kloop & 0.6237 & 0.5846 & 0.6684 \\
\hfil 18 & nlpseattle & 0.6201 & 0.6332 & 0.6075 \\
\hfil 19 & gw2018 & 0.6038 & 0.5158 & 0.7280 \\
\hfil 20 & MindCoders & 0.5858 & 0.5264 & 0.6603 \\
\hfil 21 & NSIT & 0.5794 & 0.6614 & 0.5155 \\
\hfil 22 & Summer2019 & 0.5567 & 0.6724 & 0.4749 \\
\hfil 23 & Antiganda & 0.5490 & 0.6609 & 0.4695 \\
\hfil 24 & Cojo & 0.5472 & 0.6692 & 0.4627 \\
\hfil 25 & \underline{Baseline} & 0.4734 & 0.4437 & 0.5074 \\
\hfil 26 & gudetama & 0.4734 & 0.4437 & 0.5074 \\
\hfil 27 & test & 0.4734 & 0.4437 & 0.5074 \\
\hfil 28 & Visionators & 0.4410 & 0.5909 & 0.3518 \\
\hfil 29 & MaLaHITJuniors & 0.3075 & 0.4694 & 0.2286 \\
\bottomrule
	\end{tabular}
	\caption{Results for the SLC task on the development set at the end of phase 1 (see Section~\ref{sec:setup}).\label{tab:slcdev}}
\end{table}

Team \textbf{CUNLP}~\cite{Alhindi:19} considered two approaches for the sentence-level task. The first approach was based on fine-tuning BERT. The second approach complemented the fine-tuned BERT approach by feeding its decision into a logistic regressor, together with features from the Linguistic Inquiry and Word Count (LIWC)\footnote{\url{http://liwc.wpengine.com/}} lexicon and punctuation-derived features. Similarly to \citet{Gupta:19}, for the fragment-level problem they used a Bi-LSTM-CRF architecture, combining both character- and word-level embeddings.

Team \textbf{ProperGander}~\cite{Madabushi:19} also used BERT, but they paid special attention to the imbalance of the data, as well as to the differences between training and testing. They showed that augmenting the training data by oversampling yielded improvements when testing on data that is temporally far from the training (by increasing recall). In order to deal with the imbalance, they performed cost-sensitive classification, i.e.,~the errors on the smaller positive class were more costly. For the fragment-level classification, inspired by named entity recognition, they used a model based on BERT using Continuous Random Field stacked on top of an LSTM.

\begin{table}[t]
	\centering
	\small
	\begin{tabular}{c@{\hspace{1mm}}p{2cm}ccc}
	\toprule	
	\multicolumn{5}{c}{\textbf{FLC Task: Test Set (Official Results)}}\\
	\textbf{Rank} & \textbf{Team} & \textbf{F$_1$} & \textbf{Precision} & \textbf{Recall} \\
\midrule
\hfil 1 & \textbf{newspeak} & \textbf{0.2488} & 0.2862 & 0.2200\\
\hfil 2 & Antiganda & 0.2267 & 0.2882 & 0.1868\\
\hfil 3 & MIC-CIS & 0.1998 & 0.2234 & 0.1808\\
\hfil 4 & Stalin & 0.1453 & 0.1920 & 0.1169\\
\hfil 5 & TeamOne & 0.1311 & 0.3234 & 0.0822\\
\hfil 6 & aschern & 0.1090 & 0.0715 & 0.2294\\
\hfil 7 & ProperGander & 0.0989 & 0.0651 & 0.2056\\
\hfil 8 & Sberiboba & 0.0450 & 0.2974 & 0.0243\\
\hfil 9 & BananasInPajamas & 0.0095 & 0.0095 & 0.0095\\
\hfil 10 & JUSTDeep & 0.0011 & 0.0155 & 0.0006\\
\hfil 11 & \underline{Baseline} & 0.0000 & 0.0116 & 0.0000\\
\hfil 12 & MindCoders & 0.0000 & 0.0000 & 0.0000\\
\hfil 13 & SU & 0.0000 & 0.0000 & 0.0000\\
\bottomrule
\end{tabular}
\caption{\label{tab:flc:results}Official test results for the FLC task.}
\end{table}

\section{Evaluation Results}
\label{sec:results}

The results on the test set for the SLC task are shown in Table~\ref{tab:slctest}, while Table~\ref{tab:slcdev} presents the results on the development set at the end of phase 1 (cf.\ Section~\ref{sec:setup}).\footnote{Upon request from the participants, we reopened the submission system for the development set for both tasks after the end of \mbox{phase 2}; therefore, Tables~\ref{tab:slcdev} and~\ref{tab:flc:dev} might not be up to date with respect to the online leaderboard.} 
The general decrease of the F$_1$ values between the development and the test set could indicate that systems tend to overfit on the development set. Indeed, the winning team \textbf{ltuorp} chose the parameters of their system both on the development set and on a subset of the training set in order to improve the robustness of their system. 

Tables~\ref{tab:flc:results} and~\ref{tab:flc:dev} report the results on the test and on the development sets for the FLC task. For this task, the results tend to be more stable across the two sets. Indeed, team \textbf{newspeak} managed to almost keep the same difference in performance with respect to team \textbf{Antiganda}. Note that team \textbf{MIC-CIS} managed to reach the third position despite never having submitted a run on the development set.


\section{Conclusion and Further Work}
\label{sec:conclusion}

We have described the NLP4IF@EMNLP-IJCNLP 2019 shared task on fine-grained propaganda identification. 
We received 25 and 12 submissions on the test set for the sentence-level classification and the fragment-level classification tasks, respectively. Overall, the sentence-level task was easier and most submitted systems managed to outperform the baseline. The fragment-level task proved to be much more challenging, with lower absolute scores, but most teams still managed to outperform the baseline.

We plan to make the schema and the dataset publicly available to be used beyond NLP4IF. We hope that the corpus would raise interest outside of the community of researchers studying propaganda: the techniques related to fallacies and the ones relying on emotions might provide a novel setting for researchers interested in Argumentation and Sentiment Analysis.

As a kind of advertisement, Task 11 at SemEval 2020\footnote{\url{http://propaganda.qcri.org/semeval2020-task11/}} is a follow up of this shared task. It features two complimentary tasks:
\begin{description}
 \item[Task~1] Given a free-text article, identify the propagandist text spans.
\item[Task~2] Given a text span already flagged as propagandist and its context, identify the specific propaganda technique it contains. 
\end{description}

This setting would allow participants to focus their efforts on binary sequence labeling for Task~1 and on multi-class classification for Task~2.

\begin{table}[t]
	\centering
	\small
	\begin{tabular}{p{0.4cm}p{2cm}ccc}
	\toprule	
	\multicolumn{5}{c}{\textbf{FLC Task: Development Set}}\\
\textbf{Rank} & \textbf{Team} & \textbf{F$_1$} & \textbf{Precision} & \textbf{Recall} \\
\midrule
\hfil 1 & newspeak & 0.2422 & 0.2893 & 0.2084\\
\hfil 2 & Antiganda & 0.2165 & 0.2266 & 0.2072\\
\hfil 3 & Stalin & 0.1687 & 0.2312 & 0.1328\\
\hfil 4 & ProperGander & 0.1453 & 0.1163 & 0.1934\\
\hfil 5 & KS & 0.1369 & 0.2912 & 0.0895\\
\hfil 6 & TeamOne & 0.1222 & 0.3651 & 0.0734\\
\hfil 7 & aschern & 0.1010 & 0.0684 & 0.1928\\
\hfil 8 & gudetama & 0.0517 & 0.0313 & 0.1479\\
\hfil 9 & AMT & 0.0265 & 0.2046 & 0.0142\\
\hfil 10 & esi & 0.0222 & 0.0308 & 0.0173\\
\hfil 11 & ltuorp & 0.0054 & 0.0036 & 0.0107\\
\hfil 12 & \underline{Baseline} & 0.0015 & 0.0136 & 0.0008\\
\hfil 13 & CAUnLP & 0.0015 & 0.0136 & 0.0008\\
\hfil 14 & JUSTDeep & 0.0010 & 0.0403 & 0.0005\\
\bottomrule
\end{tabular}
\caption{\label{tab:flc:dev}Results for FLC tasl on the development set. The values refer to the end of phase 1 (see section~\ref{sec:setup})}
\end{table}

\section*{Acknowledgments}

This research is part of the Propaganda Analysis Project,\footnote{\url{http://propaganda.qcri.org}} which is framed within the Tanbih project.\footnote{\url{http://tanbih.qcri.org}} 
The Tanbih project aims to limit the effect of ``fake news'', propaganda, and media bias by making users aware of what they are reading, thus promoting media literacy and critical thinking, which is arguably the best way to address disinformation and ``fake news.'' The project is developed in collaboration between the Qatar Computing Research Institute (QCRI), HBKU and the MIT Computer Science and Artificial Intelligence Laboratory (CSAIL). 

The corpus for the task was annotated by A Data Pro,\footnote{\url{http://www.aiidatapro.com}} a company that performs high-quality manual annotations.

\bibliography{biblio}
\bibliographystyle{acl_natbib}
\end{document}